\documentclass[runningheads]{llncs}

\usepackage{eccv}

\usepackage{eccvabbrv}

\usepackage{graphicx}
\usepackage{booktabs}
\usepackage{multirow}
\usepackage{enumitem}
\usepackage{amssymb}
\usepackage{pifont}
\usepackage[accsupp]{axessibility}  

\usepackage{hyperref}

\usepackage{orcidlink}

\begin{document}


\title{TAR: Temporal Anchor-Constrained Reasoning for Video Temporal Grounding} 

\titlerunning{TAR}

\author{Chaohong Guo\inst{1}\orcidlink{0009-0009-1452-2524}$^{*}$ \and
Xun Mo\inst{1}\orcidlink{0009-0009-3377-2107}$^{*}$ \and
Yongwei Nie\inst{1}\orcidlink{0000-0002-8922-3205}$^{\dagger}$ \and \\
Fei Ma\inst{2}\orcidlink{0009-0002-5388-9125}$^{\dagger}$ \and
Xuemiao Xu\inst{1}\orcidlink{0000-0002-8006-3663} \and
Chengjiang Long\inst{3}\orcidlink{0000-0003-1584-7290}$^{\ddagger}$}

\authorrunning{C.~Guo, X.~Mo et al.}

\institute{South China University of Technology \and
Guangdong Laboratory of Artificial Intelligence and Digital Economy (SZ) \and
Bytedance Inc.\\
\email{cguo5104@gmail.com} \quad \email{nieyongwei@scut.edu.cn}}

\maketitle

\begingroup
\renewcommand{\thefootnote}{}
\footnotetext{
$^{*}$ Equal contribution.\\
$^{\dagger}$ Corresponding author.\\
$^{\ddagger}$  This work was done when Dr. Chengjiang Long was with Meta.
}
\endgroup

\begin{abstract}
Video Temporal Grounding (VTG) aims to localize specific video segments corresponding to natural language queries. While recent Large Vision-Language Models (LVLMs) employ Reinforcement Learning to generate Chains-of-Thought (CoT), they typically rely solely on outcome-based supervision. Consequently, this often leads to hallucinations, where the reasoning process becomes disconnected from the visual content and the final prediction. Existing attempts to mitigate this by relying on external supervision from larger models or separate reward models are computationally expensive and prone to rigid patterns. To address these challenges, we propose \textbf{TAR} (\textbf{T}emporal \textbf{A}nchor-Constrained \textbf{R}easoning), a framework that introduces the temporal anchor (T-anchor) as a transparent and auditable checkpoint mechanism. T-anchor enforces progressive refinement within the CoT, compelling the model to continuously ground its intermediate thoughts in visual evidence and iteratively calibrate temporal predictions, thereby significantly enhancing the faithfulness and autonomy of the reasoning process and final accuracy. Furthermore, we introduce a bootstrapping paradigm that automatically harvests high-quality CoT data using only a standard 7B model, eliminating the dependency on ultra-large models. Extensive experiments demonstrate that TAR achieves state-of-the-art performance and generates faithful, autonomous, and progressively refined reasoning traces.
Code is available at \url{https://github.com/chaohongguo/TAR}.
\end{abstract}

\section{Introduction}

Video Temporal Grounding (VTG) is a fundamental task in video understanding that requires models to localize specific segments within untrimmed videos corresponding to a given natural language description. Envisioning VTG technology empowering AI assistants to pinpoint specific moments (such as ``a child entering the kitchen'') from hours of footage, this capability is critical for real-world applications including video surveillance, intelligent video retrieval, and human-computer interaction.

Existing VTG methods generally fall into three categories. First, Vision-Language Pre-training (VLP) methods extract features using pre-trained encoders and employ specific modules for fusion and localization. However, the disjoint training of feature extraction and localization often leads to error accumulation. Second, methods based on Supervised Fine-Tuning (SFT) of Large Vision-Language Models (LVLMs) typically rely on next-token prediction (cross-entropy loss). This objective aligns poorly with the Intersection-over-Union (IoU) metric, and since these models usually output direct answers without reasoning chains, they lack interpretability. Third, recent approaches utilize Reinforcement Learning (RL) to encourage models to generate a Chain-of-Thought (CoT) denoted as \texttt{<think>...</think>} before producing the final \texttt{<answer>}.

\begin{figure}[t]
\centering
\includegraphics[width=1\columnwidth]{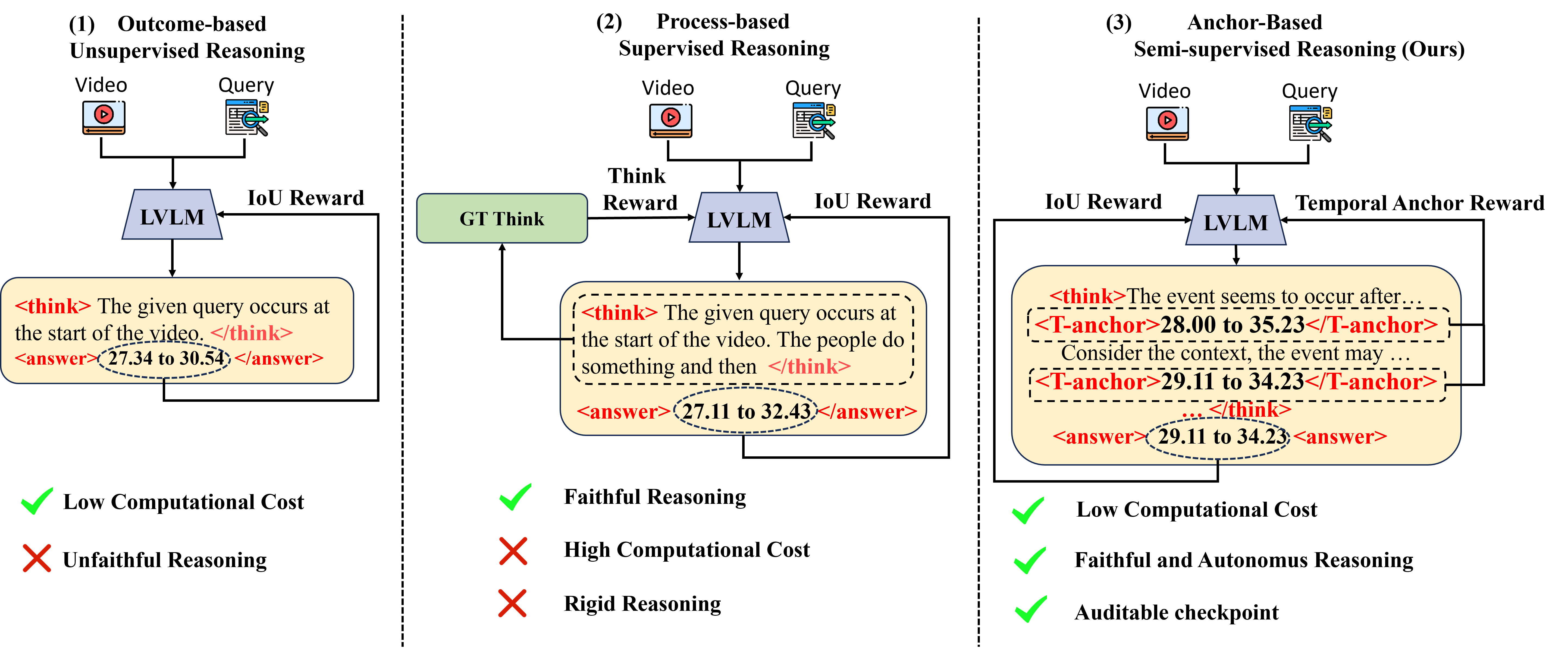}
\caption{\textbf{Comparison of reasoning paradigms.} (1) \textbf{Outcome-based Unsupervised Reasoning} relies solely on final outcome supervision, offering low computational cost but leading to unfaithful reasoning. (2) \textbf{Process-based Supervised Reasoning} depends on step-by-step supervision (GT Think), resulting in faithful reasoning but suffering from high computational costs and rigid patterns. (3) \textbf{Anchor-Based Semi-supervised Reasoning (Ours)} introduces auditable temporal anchors (\texttt{<T-anchor>}) as checkpoints. By utilizing a Temporal Anchor Reward, our method achieves faithful and autonomous reasoning while maintaining a low computational cost.}
\label{fig:intro}
\end{figure}

A core challenge remains: ensuring the reasoning process is tightly grounded in video content to effectively enhance localization accuracy. As illustrated in Figure~\ref{fig:intro} (1), methods~\cite{li2025reinforcement,yan2025videochat} like Time-R1~\cite{wang2025time} rely solely on final prediction constraints driven by an IoU Reward. We term this paradigm \textit{Outcome-based Unsupervised Reasoning}. While it boasts low computational cost, these methods often result in the ``right answer for the wrong reason'' phenomenon, where the generated reasoning suffers from hallucinations or generic descriptions detached from specific visual cues, leading to unfaithful reasoning. In contrast, as shown in Figure~\ref{fig:intro} (2), other approaches like Video-VER~\cite{luo2025thinking} which we categorize as \textit{Process-based Supervised Reasoning} typically rely on an external LVLM to generate ground truth thoughts (GT Think) and explicitly score the reasoning process via a reward to think process. While this achieves faithful reasoning, excessive reliance on such explicit supervision leads to high computational cost and rigid reasoning output patterns that lack autonomy, reducing the model to a template-matcher that merely parrots predefined structures (e.g., "The person is doing [Action]") to satisfy the external scorer.

To address these challenges, we propose \textbf{TAR} (\textbf{T}emporal \textbf{A}nchor-Constrained \textbf{R}easoning), as illustrated in Figure~\ref{fig:intro} (3), which serves as an efficient semi-supervised reasoning paradigm. The core innovation of TAR lies in embedding explicit \texttt{<T-anchor>...</T-anchor>} tags within the reasoning trace to serve as auditable checkpoints. Instead of relying on heavy external models to evaluate the intermediate steps, we design a progressive refinement strategy driven by a new reward mechanism. Specifically, alongside the standard IoU Reward for the final \texttt{<answer>}, we introduce a novel Temporal Anchor Reward applied directly to the intermediate \texttt{<T-anchor>} tags. During generation, the model first proposes a coarse interval at an initial anchor. It is then compelled to re-think and ground its subsequent reasoning in visual evidence, producing a more precise interval at the next anchor. By rewarding the model strictly for these incremental accuracy gains, our mechanism ensures that the reasoning process actively aligns with the visual content. Consequently, TAR achieves faithful and highly autonomous reasoning with minimal computational overhead.

Another significant challenge is that base models often struggle to strictly follow formatting instructions for T-anchors. Experiments show that even with explicit prompting, up to $76\%$ of samples fail to generate valid T-anchors (e.g., unclosed tags), rendering anchor-based constraints ineffective and hindering training efficiency. Instead of distilling from larger models, we propose a bootstrapping strategy: performing lightweight RL directly on the 7B model to harvest $30k$ well-formatted reasoning chains as seed data for high-quality training.

Our main contributions are as follows:
\begin{itemize}[label=\textbullet]
    \item We propose the \textbf{TAR} framework, which embeds verifiable T-anchors into the reasoning chain to encourage the model to progressively refine its temporal predictions and continuously ground its intermediate thoughts in visual evidence. This effectively bridges the gap between the reasoning process and localization results, enhancing both faithfulness and accuracy.
    \item We introduce a resource-efficient bootstrapping data generation paradigm, eliminating the dependency on ultra-large models for generating high-quality reasoning chains.
    \item Experimental results demonstrate the superiority of TAR, establishing a new performance benchmark on the Charades-STA dataset with $61.1$ mIoU and $50.2$ R1@0.7.
\end{itemize}

\section{Related Works}

\textbf{Temporal Video Grounding.} Temporal Video Grounding (TVG) \cite{gao2017tall,krishna2017dense} aims to precisely locate video segments corresponding to text queries. Early methods mainly rely on Vision-Language Pre-training (VLP) models \cite{zhang2021ms2dtan,lei2021detecting,lin2023univtg,cao2025flashvtg}, while subsequent approaches leverage large language models (LLMs) with fine-tuned visual encoders or open-source large vision-language models \cite{qu2024chatvtg,huang2024vtimellm,zeng2025timesuite,ren2024timechat,guo2026t2sgrid}, though both lack interpretability and temporal awareness. Recently, reasoning and reinforcement learning (RL) have been introduced into LLMs \cite{jin2025search,wu2025mmsearch}, with Time-R1 \cite{wang2025time} achieving state-of-the-art grounding performance through a reasoning-driven LVLM and VideoChat-R1 \cite{li2025videochat} extending to broader spatio-temporal tasks, yet these RL-based methods still lack explicit constraints on the reasoning process, which our method addresses.

\noindent\textbf{Reasoning in Large Vision-Language Models.} 
Building on the success of reasoning models like OpenAI o1 and DeepSeek-R1 \cite{guo2025deepseek} in symbolic domains, recent efforts have extended Chain-of-Thought (CoT) capabilities to the visual domain \cite{zhang2025chain, zhang2025thyme}. In the context of video understanding, existing reasoning-based methods generally follow two paradigms: (1) \textit{Outcome-based Unsupervised Reasoning}: Recent methods such as TVG-R1~\cite{chen2025datasets}, Video-R1~\cite{feng2025video}, and Time-R1~\cite{wang2025time} rely solely on final task-specific metrics (e.g., IoU) as rewards. While computationally efficient, this paradigm lacks intermediate supervision, often resulting in a disconnect between the reasoning process and the final answer. Consequently, the thinkprocess may suffer from hallucinations, yielding "the right answer for the wrong reason."
(2) \textit{Process-based Supervised Reasoning}: Some approaches~\cite{wang2025skywork, luo2025thinking} incorporate additional reward models or external LLMs to evaluate the reasoning process itself. However, these model-based rewards are often computationally expensive and can lead to rigid, template-like reasoning patterns. To strike a balance, our TAR framework introduces a novel paradigm: \textit{Anchor-based Constrained Reasoning}. Unlike unsupervised methods, we enforce intermediate verification using rule-based temporal anchors; unlike fully supervised methods, we maintain semantic autonomy by only constraining the anchor format rather than the entire linguistic content. This effectively combines the efficiency of rule-based rewards with the faithfulness of process-level supervision.

\section{Methodology}

Given an untrimmed video $V$ and a query $Q$, Temporal Video Grounding (TVG) aims to predict the target segment $T = (t_s, t_e)$. We utilize a Large Vision-Language Model (LVLM) parameterized by $\theta$ to address this task. Instead of directly regressing the boundaries, we explicitly model the reasoning process by prompting the LVLM to generate a reasoning chain $R$, which includes a sequence of intermediate anchors $\mathcal{A} = \{A_1, ..., A_K\}$, followed by the final prediction $A_{final}$. The joint generation process is expressed as:
\begin{equation}
P_\theta(R, A_{final} | V, Q) = \prod_{t} P_\theta(y_t | y_{<t}, V, Q)
\end{equation}
Here, the reasoning trace $R$ is delimited by \texttt{<think>} tags, and each temporal anchor $A_i=(t_s^i,t_e^i)$ is enclosed in \texttt{<T-anchor>} tags.

First, Sec.~\ref{subsec:theory} provides the  intuition behind the effectiveness of the anchor mechanism. Next, Sec.~\ref{subsec:tartvg_reward} introduces the reward function for TAR. Finally, Sec.~\ref{subsec:data_collect} details the bootstrapping data collection method used to generate the reasoning chains.

\subsection{Intuition: Why Temporal Anchors Work.}
\label{subsec:theory}

Instead of relying on rigid token-level supervision or unconstrained generation, TAR leverages the inherent autoregressive nature of Large Vision-Language Models (LVLMs) to balance reasoning faithfulness and semantic autonomy. Purely unsupervised Chain-of-Thought (CoT) often suffers from ``textual inertia'' where the model pays more attention to its newly generated text than the visual input, leading to ungrounded hallucinations. The anchor mechanism acts as a structural interruption to break this inertia. Because our reward strictly penalizes any drop in localization accuracy, the model cannot simply guess the next timestamp based on text alone. It is forced to shift its attention back to the raw video tokens to find concrete evidence. Thus, each T-anchor serves as a ``visual grounding checkpoint,'' ensuring the reasoning is deeply faithful to the actual video. At the same time, TAR preserves the model's semantic autonomy. Unlike supervised methods that reduce the model to a template-matcher forced to output fixed patterns (e.g., ``The person is doing [Action]''), TAR only constrains the format of the numerical anchors. The natural language reasoning between these checkpoints remains completely free-form. This allows the model to autonomously explore and describe the video content without being restricted by rigid pseudo-labels.

In essence, this decoupled design of strict temporal anchor verification and free-form text generation empowers the model to achieve highly accurate localization while maintaining both robust reasoning faithfulness and genuine semantic autonomy.

\begin{figure*}[t!]
\centering
\includegraphics[width=1.0\textwidth]{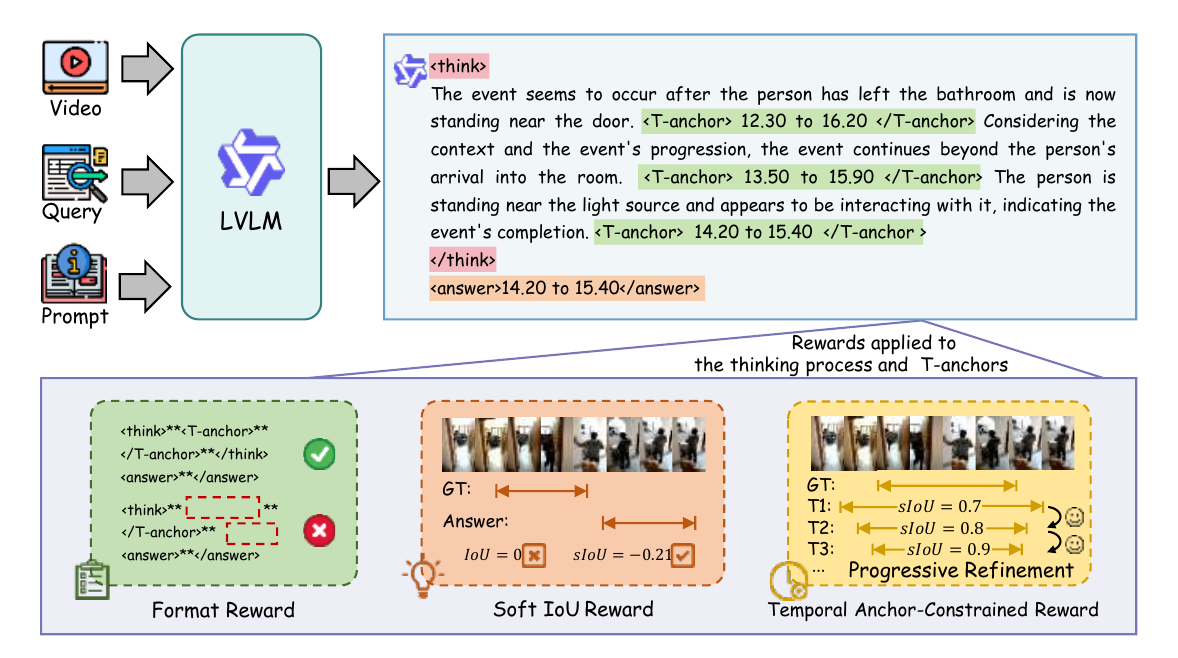}
\caption{Overview of our proposed Temporal Anchor-constrained Reasoning for Temporal Video Grounding (TAR) method. We apply a format reward, a soft IoU reward, and a temporal anchor-constrained reward to the think, answer and T-anchor tags generated by the LVLM.} 
\label{fig:gch-mx-fig2.png}
\end{figure*}

\subsection{The TAR Framework}
\label{subsec:tartvg_reward}
Guided by the analysis in Sec.~\ref{subsec:theory}, we introduce the \textbf{TAR} framework. We employ Reinforcement Learning (specifically GRPO~\cite{guo2025deepseek}) to optimize the model, using the proposed T-anchors as auditable checkpoints. We design a composite reward function $R(o)$ consisting of three components:

\noindent\textbf{(1) Format Reward ($r_{Format}$).}
We prompt the model to output in the following structure: 
{\small \texttt{<think>...<T-anchor> \\ 
...</T-anchor>...<T-anchor>...</T-anchor>... </think>  <answer>...</answer>}}.

In our \texttt{<think>} block, the model first provides an initial analysis of the video, then outputs a preliminary answer in the first \texttt{<T-anchor>} block. It then continues to analyze this initial answer in combination with its earlier reasoning and the video content, producing a refined or corrected answer in the second \texttt{<T-anchor>} block, and so on until it wants to stop. In the \texttt{<answer>} block, the model provides its final answer, which we set as the same in the last T-anchor tag in the think block. We use regular expressions to verify whether the model's output conforms to this required format. If the format is correct, we give a score of $\alpha$ to the model, otherwise 0:

\begin{equation}
    r_{\mathrm{Format}}(o) = 
            \begin{cases}
                 0, \mathrm{if}\ o\mathrm{\ does\ not\ match\ format}   \\
                 \alpha, \mathrm{if}\ o\mathrm{\ matches \ format}
            \end{cases}
\end{equation}
where we use $\alpha=3$, see Supplementary Material
for ablation study.

\noindent\textbf{(2) Soft IoU Reward ($r_{sIoU}$).}
In previous work, they usually adopt the following IoU reward: 
\begin{equation}
r_{\mathrm{IoU}}(o) = \frac{\max\left(0, \min(t_e, \hat{t}_e) - \max(t_s, \hat{t}_s)\right)}{\max(t_e, \hat{t}_e) - \min(t_s, \hat{t}_s)}.
\end{equation}
In this paper, we propose a slightly different soft IoU to compute the reward for the model's output:
\begin{equation}
        r_{\mathrm{sIoU}}(o) = \frac{\min(t_e, \hat{t}_e) - \max(t_s, \hat{t}_s)}{\max(t_e, \hat{t}_e) - \min(t_s, \hat{t}_s)}.
\end{equation}
The only difference is that the numerator has no lower bound of zero. Soft IoU can measure the distance between the predicted and ground-truth segments even when there is no overlap which the previous IoU cannot achieve, enabling stable training. For example, in the early stages of training, if the model predicts a segment from 15.4s to 18.0s while the ground truth is from 3.4s to 12.2s, soft IoU yields a score of -0.21, whereas standard IoU gives 0. As a result, the standard IoU is not sensitive to the degree of misalignment in the predictions, leading to slower convergence.

\noindent\textbf{(3) Temporal Anchor-Constrained Reward ($r_{TAR}$).}
Besides the standard format and soft IoU reward functions, we further design a temporal anchor-constrained reward to apply explicit constraints to the generated anchors. First, we identify format-correct \texttt{<T-anchor>} tags within the reasoning content and extract the time durations in-between them. Let $o$ be the model output, $s$ be the target number of \texttt{<T-anchor>}  tags, and $\hat{s}$ be the actual number generated by the model. The $r_{\text{TAR}}$ consists of three sub-terms:

\textbf{Weighted Accuracy:} We assign higher weights to later anchors to encourage convergence and enforce them to be more accurate towards the final prediction. We compute the weighted sum of the soft IoUs between the predicted time ranges and the ground truth:
\begin{equation}
    \text{TAR}_\mathrm{sIoU}(o) = \sum_{i=1}^{\hat{s}} i \cdot \text{sIoU}_{i}(o),
\end{equation}
where $\text{sIoU}_{i}(o)$ computes the soft IoU score for the $i$-th anchor of output $o$, and $i$ acts as the increasing weight.

\textbf{Progressive Refinement:} We explicitly reward the model only when it refines its prediction, requiring that later temporal anchors be more accurate than earlier ones. The progressive refinement step for the $i$-th anchor is formalized as:
\begin{equation}
    \delta_i(o) = 
    \begin{cases} 
        1 & \text{if } \text{sIoU}_{i}(o) > \text{sIoU}_{i-1}(o), \\
        -1 & \text{otherwise.}
    \end{cases}
\end{equation}
Based on this, the total progressive refinement reward is defined as:
\begin{equation}
    \text{TAR}_\mathrm{refine}(o) = \sum_{i=2}^{\hat{s}} \delta_i(o).
\end{equation}

\textbf{Length Penalty:} If the size of $\hat{s}$ is unbounded, the model might engage in reward hacking by generating infinite anchors to accumulate $\text{TAR}_\mathrm{sIoU}$, which increases the difficulty of progressive refinement and can ultimately degrade the model's training performance. To prevent $\hat{s}$ from deviating from the target count $s$, we impose a quadratic penalty:
\begin{equation}
    \text{TAR}_\mathrm{num}(o) = (\hat{s} - s)^2.
\end{equation}
Finally, our complete temporal anchor-constrained reward is formulated as:
\begin{equation}
    r_{\text{TAR}}(o) = \text{TAR}_\mathrm{sIoU}(o) + \beta \cdot \text{TAR}_\mathrm{refine}(o) - \gamma \cdot \text{TAR}_\mathrm{num}(o),
\end{equation}
where $\beta=1$ and $\gamma=5$ are hyperparameters balancing the refinement incentive and length constraint. 

The overall reward of the whole reinforcement learning is formulated as:
\begin{equation}
R(o) = r_{\mathrm{Format}}(o) + r_{\text{sIoU}}(o) + r_{\text{TAR}}(o).
\label{eqa:all}
\end{equation}

\begin{figure*}[t!]
\centering
\includegraphics[width=1.0\textwidth]{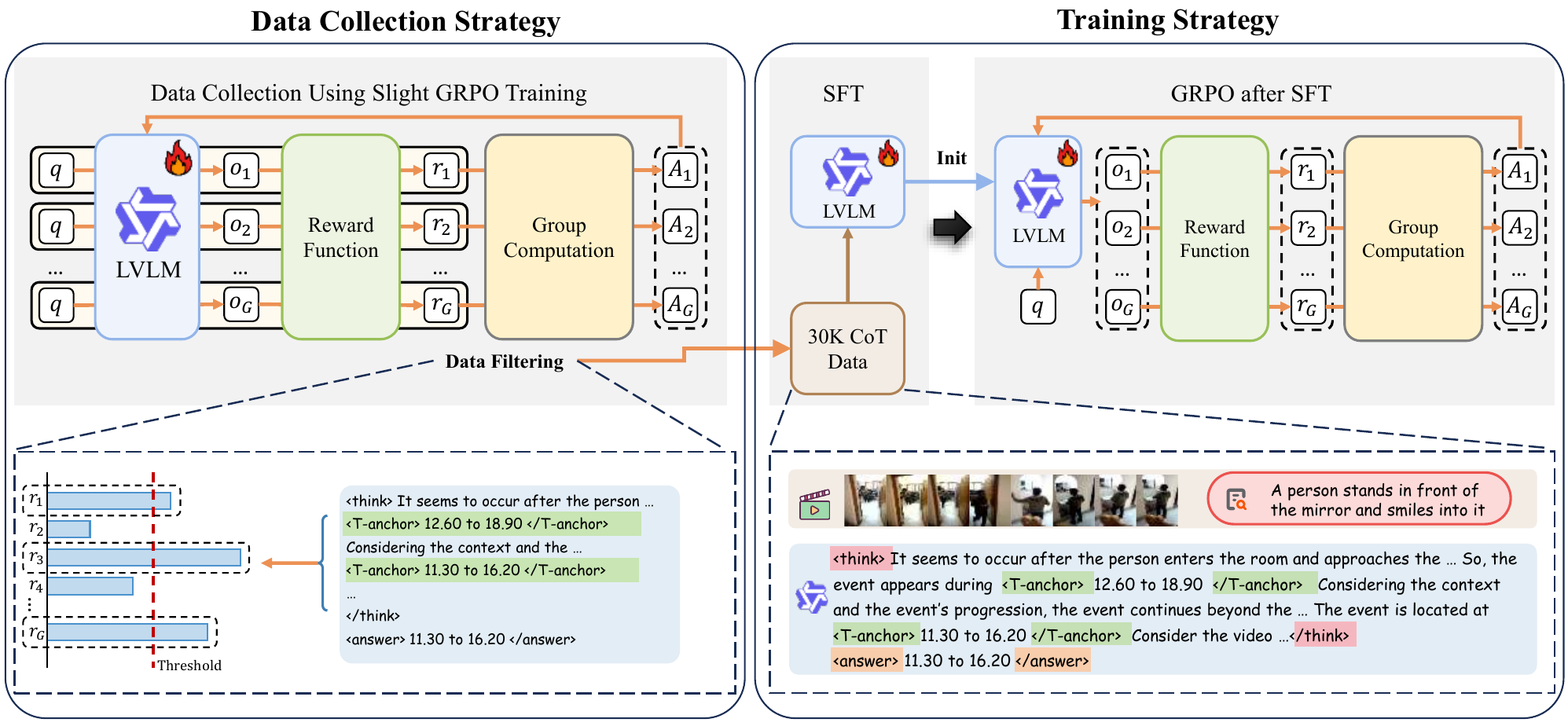}
\caption{\textbf{Data collection strategy (left) and training strategy (right).} We first employ a lightweight GRPO training to generate reasoning data, which is then filtered to obtain high-quality samples. Using this data, we perform Supervised Fine-Tuning (SFT) on the base Qwen2.5-VL model for a cold start, followed by final GRPO training on the SFT model.}
\label{fig:data_gen.png}
\end{figure*}

\subsection{Bootstrapping Strategy for Data Efficiency}
\label{subsec:data_collect}

Training the proposed model directly is non-trivial. Despite extensive experimentation with various prompts to encourage the model to output \texttt{<T-anchor>} tags, it consistently struggles to generate them reliably. This is partly because the baseline model we use (i.e., Qwen2.5-VL-3B/7B) has limited capability in fully comprehending complex temporal reasoning prompts. When no \texttt{<T-anchor>} tags are produced, the temporal anchor-constrained reward cannot be computed, which prevents effective training.

To circumvent this, we introduce a Self-Bootstrapping Data Collection strategy via Slight GRPO Training (Fig. \ref{fig:data_gen.png}). Unlike conventional knowledge distillation methods that incur prohibitive financial and computational costs by relying on massive, proprietary closed-source models (e.g., GPT-4o or Gemini 1.5 Pro) for data generation, our approach is highly cost-effective. It leverages the baseline model's own exploratory capabilities to synthesize reasoning trajectories. Given a video $V$, prompt $p$, and query $q$ sourced the training splits of dataset, the GRPO algorithm generates a diverse group of candidates $o$ and computes their respective rewards $r$ to optimize the policy. 

To ensure the quality of the bootstrapped data, we apply a rigorous, deterministic filtering pipeline. From an initial pool of 186K raw exploratory outputs generated from the training set, we distilled 30K high-quality reasoning samples using precise criteria (as illustrated in Fig. \ref{fig:data_gen.png}). Specifically, each retained trajectory must achieve a high total reward, which requires the reasoning trajectory to contain a minimum of \texttt{<T-anchor>} tags numbers. Crucially, each sequential anchor must demonstrate progressively increasing localization accuracy against the ground truth, with the specific accuracy thresholds (e.g., sIoU). This automated pipeline effectively bootstraps a high-quality supervision signal from extensive exploratory trajectories, completely eliminating the need for manual intervention or costly external large-model distillation.

\section{Experiments}

\subsection{Experimental Setup}
\label{subsec:exp_setup}
\textbf{Benchmarks.} We evaluate our model on four video grounding datasets: Charades-STA~\cite{gao2017tall} (6,672 indoor activity videos, 30.6s avg, 12,408 train/3,720 test clip-query pairs), QVHighlights~\cite{lei2021detecting} (4,855 train/1,020 val video-text pairs that contain only a single segment), ActivityNet-Caption~\cite{caba2015activitynet} (37,421 train/17,505 val/17,031 test samples), and TVGBench~\cite{wang2025time} (800 samples). The last one aggregates part test sets from ActivityNet-Caption, Charades-STA, HiREST~\cite{zala2023hierarchical}, EgoNLQ~\cite{grauman2022ego4d} and TaCoS~\cite{regneri2013grounding} as a comprehensive benchmark, evaluating 11 query types with balanced video-query distribution for fair assessment.

\begin{table*}[t]
\centering

\caption{
Performance of temporal video grounding on Charades-STA, QVHighlights.
Methods marked with $^*$ are trained with additional TVG datasets (See the Supplementary Material for their implementation and the dataset they use), while FT with \checkmark indicates that the model is fine-tuned on the corresponding Charades-STA or QVHighlights dataset. 
We compare our method against existing 3B, 7B open-source LVLMs, as well as state-of-the-art VLP models. }
\resizebox{\textwidth}{!}{
\begin{tabular}{c l | c c | cccc | cccc }
\toprule
\multirow{2}{*}{Type} & \multirow{2}{*}{Method} & \multirow{2}{*}{Size} & \multirow{2}{*}{FT} &
\multicolumn{4}{c|}{Charades-STA} & 
\multicolumn{4}{c}{QVHighlights} \\
 & & & &
{\fontsize{8.4}{10}\selectfont mIoU} &
{\fontsize{8.4}{10}\selectfont R1@0.3} & 
{\fontsize{8.4}{10}\selectfont R1@0.5} & 
{\fontsize{8.4}{10}\selectfont R1@0.7} & 
{\fontsize{8.4}{10}\selectfont mIoU} &
{\fontsize{8.4}{10}\selectfont R1@0.3} & 
{\fontsize{8.4}{10}\selectfont R1@0.5} & 
{\fontsize{8.4}{10}\selectfont R1@0.7} \\
\midrule

\multirow{6}{*}{VLP} 

& 2D-TAN~\cite{zhang20192DTAN} & - & \checkmark 
& - & 57.3 & 45.8 & 27.9 & - & - & - & -  \\

& M-DETR~\cite{lei2021detecting} & - & \checkmark 
& 45.5 & 65.8 & 52.1 & 30.6 & - & - & 53.9 & 34.8   \\ 

& UniVTG~\cite{lin2023univtg} & - & \checkmark 
& 50.1 & 70.8 & 58.1 & 35.6 & - & - & 59.7 & -  \\ 


& EaTR~\cite{jang2023knowing} & - & \checkmark 
& - & - & 68.4 & 44.9  & -  & - & 61.4 & 45.8 \\

& CG-DETR~\cite{moon2023correlation} & - & \checkmark 
& 50.1 & 70.4 & 58.4 & 36.3  & -  & - & 67.4 & 52.1  \\


& FlashVTG~\cite{cao2025flashvtg} & - & \checkmark 
& - & - & 70.32 & 49.87 & -  & - & \underline{73.1} &  \underline{57.3} \\ 
\bottomrule

\multirow{6}{*}{SFT} 

& ChatVTG~\cite{qu2024chatvtg}  & 7B & 
& - & 52.7 & 33.0 & 15.9 & - & - & - & - \\  

& TimeChat$^*$~\cite{ren2024timechat}  & 7B & \checkmark 
& - & 32.2 & 13.4 & 36.2 & - & - & - & - \\


& VTimeLLM~\cite{huang2024vtimellm}  & 7B & 
& - & 51.0 & 27.5 & 11.4 & - & - & - & - \\

 & TimeSuite~\cite{zeng2025timesuite}  & 7B & 
& - & 69.9 & 48.7 & 24.0 & - & - & - & - \\ 


 & TRACE$^*$~\cite{guo2024trace}  & 7B & \checkmark
 & - & - & 40.3 & 19.4  & - & -  & - & - \\



& TimeSuite$^*$~\cite{zeng2025timesuite} & 7B & \checkmark 
& - & 79.4 & 67.1 & 43.0 &  - & - & - & - \\ 

\bottomrule

\multirow{7}{*}{RL} 


& VideoChat-R1.5$^*$~\cite{yan2025videochat} & 3B & \checkmark 
& 50.8 & 74.9 & 58.6 & 30.9 & - & - & - & - \\

& Time-R1$^*$~\cite{wang2025time} & 3B & \checkmark 
& \underline{55.3} & \underline{79.5} & \underline{65.1} & \underline{37.5} & - & - & - & - \\

& TAR (ours) & 3B & \checkmark 
& \textbf{57.0} & \textbf{81.0} & \textbf{67.2} & \textbf{40.6} & - & - & - & - \\


& Temporal-RLT$^*$~\cite{li2025reinforcement} & 7B & \checkmark
& 57.0 & - & - & - & - & - & - & - \\

& VideoChat-R1.5$^*$~\cite{yan2025videochat} & 7B & \checkmark
& \underline{60.6} & \underline{82.8} & \underline{71.6} & 48.3 & - & - & - & - \\

& Time-R1$^*$~\cite{wang2025time} & 7B & \checkmark 
& 58.8 & \underline{82.8} & \textbf{72.2} &  \underline{50.1} & \underline{59.1} & \underline{74.1} & 66.2 & 52.7  \\ 

& TAR (ours) & 7B & \checkmark 
& \textbf{61.1} & \textbf{83.6} & 71.4 & \textbf{50.2} & \textbf{65.9} & \textbf{85.6} & \textbf{76.1 }& \textbf{58.5}  \\ 

\bottomrule
\end{tabular}}

\label{tab:comp_TVG}

\end{table*}
\begin{table}[t]
\centering
\caption{\textbf{Zero-shot} comparison on the ActivityNet-Captions and TVGBench benchmarks. All models have 7B parameters.}

\begin{tabular}{c|cccc|cccc}
\toprule
\multirow{2}{*}{Method} & \multicolumn{4}{c|}{ActivityNet-Captions} & \multicolumn{4}{c}{TVGBench} \\ 
& mIoU & R1@0.3 & R1@0.5 & R1@0.7 & mIoU & R1@0.3 & R1@0.5 & R1@0.7 \\ 
\midrule

VideoChat~\cite{li2023videochat} & 7.2 & 8.8 & 3.7 & 1.5 & - & - & - & - \\
Video-ChatGPT~\cite{maaz2023video} & 18.9 & 26.4 & 13.6 & 6.1 & - & - & - & - \\
ChatVTG~\cite{qu2024chatvtg} & 27.2 & 40.7 & 22.5 & 9.4 & - & - & - & - \\
TimeChat~\cite{ren2024timechat} & - & 36.2 & 20.2 & 9.5 & - & 22.4 & 11.9 & 5.3 \\
HawkEye~\cite{wang2024hawkeye} & - & 49.1 & 29.3 & 10.7 & - & - & - & - \\
VTime-LLM~\cite{huang2024vtimellm} & - & 44.0 & 27.8 & 14.3 & - & - & - & - \\
Videochat-Flash~\cite{li2024videochat} & - & - & - & - & - & 32.8 & 19.8 & 10.4 \\
TRACE~\cite{guo2024trace} & - & - & - & - & - & 37.0 & 25.5 & 14.6 \\
TimeSuite~\cite{zeng2025timesuite} & - & - & - & - & - & 31.1 & 18.0 & 8.9  \\


VideoChat-R1.5~\cite{yan2025videochat} & 35.5 & 52.4 & 32.3 & 16.8 & - & - & - & -  \\

Temporal-RLT~\cite{li2025reinforcement} & \underline{39.0} & - & - & - & - & - & - & -  \\

Time-R1~\cite{wang2025time} & - & \underline{58.6} & \underline{39.0} & \textbf{21.4} & \underline{29.2} & 41.8 & \underline{29.4} & \textbf{16.4}\\

TAR(Ours) & \textbf{41.1} & \textbf{61.5} & \textbf{39.8}& \underline{19.8}  & \textbf{30.6} & \textbf{42.8} & \textbf{31.1} & \underline{16.0} \\
\bottomrule
\end{tabular}

\label{tab:anet_tvg_results}
\end{table}
\textbf{Implementation details.}
We use Qwen2.5-VL-3B/7B as our base models. We first generate and filter the 30K CoT data, and then train an SFT model using this data. Subsequently, we employ GRPO to further train the model on the corresponding datasets. All experiments are conducted using 8 $\times$ NVIDIA A100 GPUs. Further details regarding the training process, including SFT and GRPO configurations along with other experimental settings, can be found in the Supplementary Material.

\subsection{Evaluation}
We compare our TAR-TVG model with state-of-the-art TVG methods, including feature-based vision-language pretraining (VLP) models, large vision-language models fine-tuned through supervised fine-tuning (SFT), and recent approaches that leverage reinforcement learning (RL) to generate chain-of-thought reasoning.

\textbf{Comparison on Charades-STA.}
As shown in Table~\ref{tab:comp_TVG}, our TAR (7B) achieves the highest mIoU of 61.1 and R1@0.3 of 83.6 on the Charades-STA benchmark. This performance is achieved using only the Charades-STA training set, surpassing methods (denoted with *) that rely on additional training data, including but not limited to YT-Temporal~\cite{yang2023vid2seq}, DiDeMo~\cite{anne2017localizing}, QuerYD~\cite{oncescu2021queryd}, InternVid~\cite{wang2023internvid}, and HowTo100M~\cite{miech2019howto100m}. For instance, in terms of R1@0.3, our model outperforms Time-R1 (82.8) trained with an additional 2.5k external samples, as well as all other types of TVG models. Our smaller 3B version of TAR surpasses most VLP-based models and reaches performance comparable to some existing 7B models.

\textbf{Comparison on QVHighlights.} On the QVHighlights dataset, our method significantly outperforms most VLP-based models and reinforcement learning-based approaches (Table.~\ref{tab:comp_TVG}). Specifically, we observe improvements of +6.8 mIoU, +9.9 R1@0.5, and +5.8 R1@0.7 compared to Time-R1, demonstrating strong generalization of our method between different TVG scenarios.

\textbf{Zero-Shot Comparison on ActivityNet-Captions and TVGBench.} In the zero-shot setting, we further evaluate our method on two challenging benchmarks: ActivityNet-Captions and TVGBench. Table~\ref{tab:anet_tvg_results} presents the results on ActivityNet-Captions, the experimental results show that our method achieves the highest R1@0.3 score (61.5) and the highest mIoU (41.1) under the zero-shot setting. Similarly, our approach achieves the highest mIoU of 30.6 on TVGBench.

\textbf{Generalization to VQA.} To evaluate the generalization capability of TAR-TVG on Video Question Answering (VQA) tasks, we allow the model to maintain its "Progressive Refinement" reasoning pattern: generating localization results as intermediate anchors within the <think> tag before producing the final categorical answer.

\begin{table}[htbp]
\centering
\caption{Comparison of Zero-shot VQA performance on MVbench~\cite{MVBench} and VideoMME~\cite{VideoMME}.}
\label{tab:vqa_generalization}
\begin{tabular}{lcc}
\toprule
\textbf{Model} & \textbf{MVBench} & \textbf{VideoMME} \\
\midrule
Qwen2.5-VL-7B  & 65.2          & 53.0 \\ 
Time-R1  & \underline{65.6}            & \underline{54.2} \\
\textbf{Ours}  & \textbf{66.1}   & \textbf{54.9}    \\
\bottomrule
\end{tabular}
\end{table}

Experimental results demonstrate that, despite being trained on TVG data, TAR-TVG has successfully acquired a "Reasoning Pattern" of continuously refining anchors during the thinking process. This ability to self-correct through intermediate anchors can be directly transferred to VQA tasks to effectively assist logical judgment, as TVG is fundamentally a subtask of video understanding. Effective temporal grounding is crucial for VQA because it helps the model focus on the most relevant moments while reducing distractions from irrelevant or redundant content. Notably, our method outperforms baseline models such as Qwen2.5-VL-7B and Time-R1 in Zero-shot VQA performance.

\subsection{Ablation Study and Discussion}
We conduct a series of ablation studies on the 7B model.

\textbf{Impact of Training Strategy.}
As shown in Table~\ref{tab:ablation_3_training}, incorporating the CoT data for supervised fine-tuning substantially enhances the model's temporal grounding ability during subsequent GRPO training. Training with reinforcement learning alone achieves only 45.9 R1@0.7, while applying chain-of-thought (CoT) supervision solely in the SFT stage yields just 21.0 R1@0.7, as SFT mainly focuses on token-level optimization and thus provides limited improvement in temporal understanding. However, when reinforcement learning is further applied to the SFT-trained model, both reasoning ability and video comprehension are strengthened, resulting in the best performance of 50.2 R1@0.7 and 61.1 mIoU.

\begin{figure*}[t]
\centering
\includegraphics[width=0.9\textwidth]{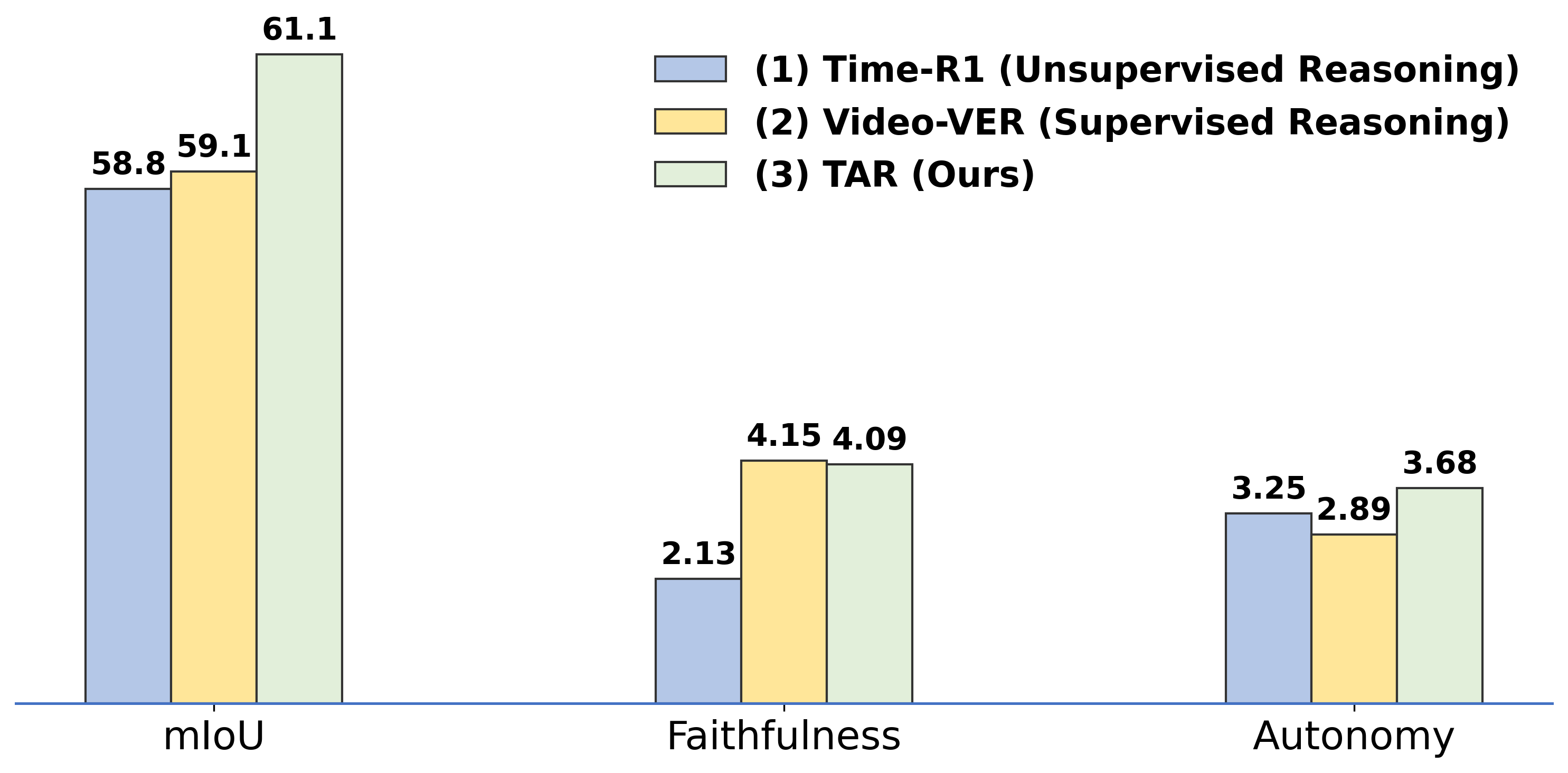}
\caption{\textbf{Quantitative comparison of reasoning trajectories and grounding performance.} Faithfulness and Autonomy are evaluated using Qwen2.5-VL-72B on a scale of 1 to 5. All methods are based on Qwen2.5-VL-7B. Our method (TAR-TVG) outperforms both Time-R1 and Video-VER, achieving the highest mIoU while maintaining robust reasoning faithfulness and enhanced autonomy.}
\label{fig:faith}

\end{figure*}

\textbf{Faithfulness and Autonomy of Reasoning Trajectories.} As illustrated in Figure~\ref{fig:faith}, we utilize Qwen2.5-VL-72B as an evaluator to score the reasoning trajectories of the three methods on a 5-point scale. Detailed evaluation protocols, including the specific prompts and scoring criteria, are provided in the supplementary material. Compared to Time-R1, which relies solely on outcome supervision, our method demonstrates superior faithfulness. Conversely, in contrast to Video-VER, which imposes strict supervision on the reasoning process, our approach exhibits enhanced autonomy. Ultimately, benefiting from our progressive refinement mechanism, our method achieves the highest mIoU. Furthermore, we provide concrete qualitative examples in the supplementary material to further illustrate the improved faithfulness and autonomy of our generated trajectories.

\begin{table}[h]

\centering
\setlength{\tabcolsep}{3pt}
\renewcommand{\arraystretch}{1.2}
\caption{
    Ablation on Training Strategy.
}
\resizebox{0.6\linewidth}{!}{
    \begin{tabular}{lcccc}
    \toprule
    Type & mIoU & R1@0.3 & R1@0.5 & R1@0.7 \\
    \midrule
    +SFT          & 40.0 & 58.8 & 41.8 & 21.0 \\
    +GRPO         & 59.3 & 81.5 & 70.1 & 45.9 \\
    +SFT + GRPO   & \textbf{61.1} & \textbf{83.6} & \textbf{71.4} & \textbf{50.2} \\
    \bottomrule
    \end{tabular}
}
\label{tab:ablation_3_training}

\end{table}
\begin{table}[h]

\centering
\setlength{\tabcolsep}{3pt}
\renewcommand{\arraystretch}{1.2}
\caption{
    Ablation on $\text{TAR}_\mathrm{sIoU}$, $\text{TAR}_\mathrm{num}$ and $\text{TAR}_\mathrm{refine}$ rewards.
}
\resizebox{0.6\linewidth}{!}{
    \begin{tabular}{ccc|ccc}
    \toprule
    $\text{TAR}_{\mathrm{sIoU}}$ & $\text{TAR}_\mathrm{refine}$ & $\text{TAR}_\mathrm{num}$ & mIoU & R1@0.5 & R1@0.7 \\
    \midrule
    \checkmark &  &  & 39.1 & 50.2 & 31.2 \\
    \checkmark & \checkmark &  & 41.1 & 51.3 & 32.4 \\
    \checkmark &  & \checkmark & 58.5 & 70.2 & 47.9 \\
    \checkmark & \checkmark & \checkmark & \textbf{61.1} & \textbf{71.4} & \textbf{50.2} \\
    \bottomrule
    \end{tabular}
}
\label{tab:ablation_tar_reward}
\end{table}

\textbf{Ablation on TAR Rewards.}
Table~\ref{tab:ablation_tar_reward} describes the ablation study on TAR rewards, involving $\text{TAR}_\mathrm{sIoU}$, $\text{TAR}_\mathrm{num}$, and $\text{TAR}_\mathrm{refine}$. In all rows, $\text{TAR}_\mathrm{sIoU}$ is always employed because it is a soft IoU constraint applied to timestamp anchors. Omitting it may generate an overly wide time span and only slightly narrow it at each step. $\text{TAR}_\mathrm{num}$ is very useful, since omitting it the model may generate an excessive number of timestamp anchors, resulting in worse performance as shown in the first two rows. With $\text{TAR}_\mathrm{sIoU}$, adding $\text{TAR}_\mathrm{num}$ achieves an R1@0.7 of 47.9. Combining all components reaches the highest 50.2, demonstrating their complementary effect.

\textbf{Performance Gain from the Anchor Mechanism.}To verify that the performance improvements stem directly from our proposed components, we compare our full TAR model (which incorporates the anchor reward detailed above) against two baselines: the standard Time-R1 (supervised solely by standard IoU) and a variant supervised by our proposed sIoU. As shown in Table~\ref{tab:ablation_anchor_mech}, under identical training datasets and configurations, our complete method consistently outperforms both baselines across all metrics. This clearly demonstrates the distinct effectiveness of both the sIoU and the anchor mechanism.

\begin{table}[b]
\centering
\caption{Ablation study on the performance gain from the anchor mechanism. All models are trained on the identical data scale.}
\label{tab:ablation_anchor_mech}
\resizebox{0.6\columnwidth}{!}{
\begin{tabular}{llcccc}
\toprule
\textbf{Model Size} & \textbf{Method} & \textbf{R1@0.3} & \textbf{R1@0.5} & \textbf{R1@0.7} & \textbf{mIoU} \\
\midrule
\multirow{3}{*}{3B} & Time-R1 & 79.5 & 65.1 & 37.5 & 55.3 \\
& Time-R1(+sIoU) & 80.1 & 65.5 & 38.1 & 56.5 \\
                    & \textbf{TAR (Ours)} & \textbf{81.0} & \textbf{67.2} & \textbf{40.6} & \textbf{57.0} \\
\midrule
\multirow{3}{*}{7B} & Time-R1 & 82.4 & 70.6 & 47.5 & 59.8 \\
& Time-R1(+sIoU) & 82.8 & 70.6 & 48.1 & 60.0 \\
                    & \textbf{TAR (Ours)} & \textbf{83.6} & \textbf{71.4} & \textbf{50.2} & \textbf{61.1} \\
\bottomrule
\end{tabular}}
\end{table}

\noindent
\begin{minipage}{0.48\textwidth}
  \textbf{Soft IoU vs Standard IoU.}
  We compared soft IoU with the standard IoU in Figure~\ref{fig:comp_siou}. In the early stages of training, soft IoU can provide informative feedback even when the predicted and ground-truth segments do not overlap, offering meaningful rewards to guide optimization. As shown in Figure~\ref{fig:comp_siou}, sIoU consistently outperforms standard IoU throughout training and leads to better final performance. 
\end{minipage}
\hfill
\begin{minipage}{0.48\textwidth}
  \centering
  \includegraphics[width=0.8\linewidth]{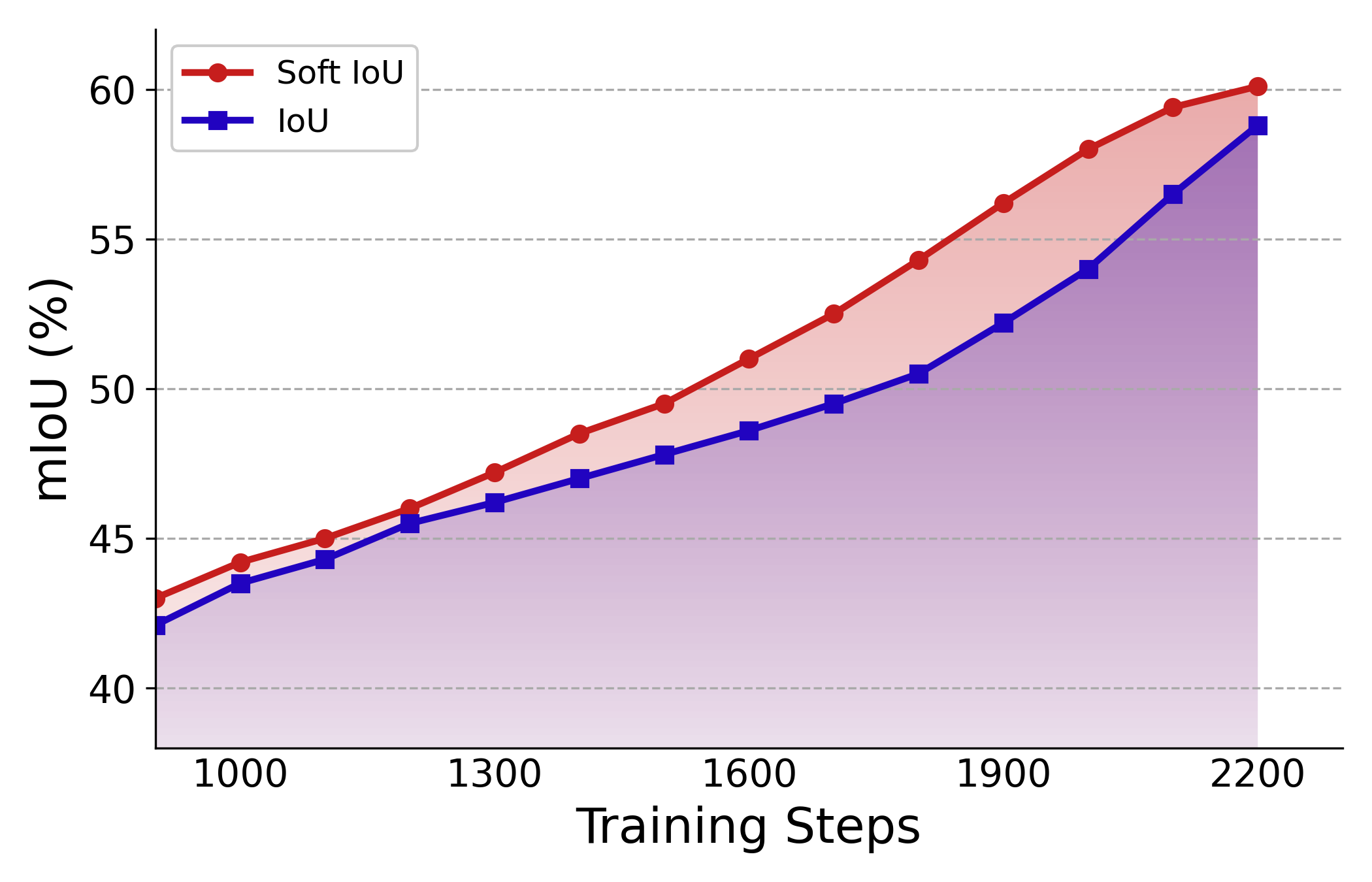}
  \captionof{figure}{Soft IoU vs Standard IoU.} 
  \label{fig:comp_siou}
\end{minipage}

\noindent
\begin{minipage}{0.48\textwidth}
  \textbf{About Thinking Length.}
  We recorded the thinking lengths of models of different sizes. 
Figure~\ref{fig:think_len_miou} show that the 3B model generally produces longer thinking sequences compared to the 7B model, while 7B model has better grouding performance than 3B. For example, when only GRPO is applied, the 3B model generates on average 11.1 (186.1-175.0) more tokens than the 7B model. 
Second, we observe that after SFT, all models show improved performance. With SFT, the thinking length of the 3B model remains nearly unchanged. For the 7B model, the thinking length decreases from 175 tokens to 158.4. Both results indicate that stronger models require less thinking to achieve better performance.
\end{minipage}
\hfill
\begin{minipage}{0.48\textwidth}
  \centering
  \includegraphics[width=0.8\linewidth]{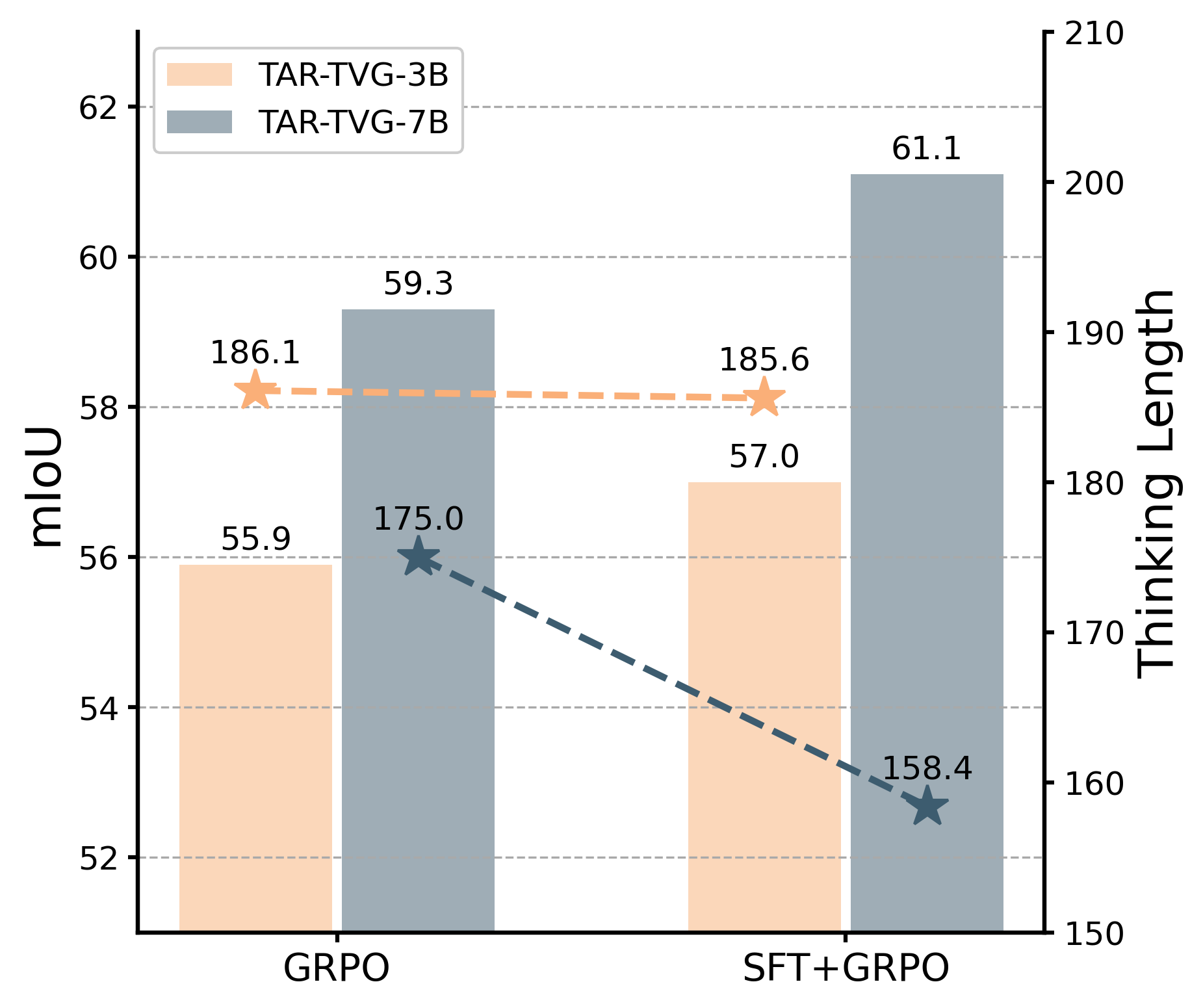}
  \captionof{figure}{The thinking length of different model size. The bar chart represents the mIoU, while the stars indicate the average number of tokens.}
  \label{fig:think_len_miou}
\end{minipage}

\textbf{Failure Analysis.} Out of 3,720 samples, we identify three major failure modes. Case~1 and Case~2 share the same quantitative behavior: the initial sIoU is 0 and remains 0 across subsequent anchors. However, they arise from different causes. In Case~1, the target action is visually similar to background activity, making it difficult to disambiguate the foreground segment. In Case~2, the target segment is extremely short, providing insufficient temporal evidence for the anchor to latch onto. Case~3 samples start near-optimal, e.g., with an sIoU of 0.87, and exhibit only minor boundary variation later, e.g., an sIoU of 0.83, with negligible impact on localization quality. Overall, 92.1\% of samples are successfully refined, indicating that incorrect initial anchors are seldom reinforced into persistent errors; the remaining cases stem from visual ambiguity, extremely short targets, or near-optimal variation.

\section{Conclusion}
In this paper, we present \textbf{TAR}, a Temporal Anchor-constrained Reasoning framework that balances reasoning faithfulness with semantic autonomy in Video Temporal Grounding. By introducing \textit{T-anchors} as auditable intermediate checkpoints, our method effectively breaks the textual inertia common in unsupervised reasoning while avoiding the rigidity of fully supervised approaches. Furthermore, we propose a resource-efficient bootstrapping paradigm that enables the model to autonomously harvest high-quality reasoning data, thereby eliminating the reliance on proprietary ultra-large models and significantly reducing computational overhead. TAR achieves state-of-the-art performance, notably reaching a record-high mIoU of 61.1 on the Charades-STA benchmark. This anchor-based constraint paradigm offers a promising direction for developing more interpretable and faithful reasoning model in the video domain.

\section*{Acknowledgments}

This work was supported by the Guangdong Basic and Applied Basic Research Foundation (2025A1515011884, 2026A1515010184), the Open Research Fund from Guangdong Laboratory of Artificial Intelligence and Digital Economy (SZ) (GML-KF-24-17), and the Research Task Assignment Project from Guangdong Laboratory of Artificial Intelligence and Digital Economy (SZ) (GML-26420007).


%
%
\bibliographystyle{splncs04}
\bibliography{main}
\end{document}